\pdfoutput=1

\documentclass[11pt]{article}

\usepackage[preprint]{acl}

\usepackage{times}
\usepackage{latexsym}
\usepackage{hyperref}
\usepackage{color, colortbl}
\definecolor{Gray}{gray}{0.9}
\usepackage{array}
\usepackage{booktabs}
\usepackage{multirow}
\usepackage{amsmath}
\usepackage{amssymb}
\usepackage{graphicx}
\usepackage{enumerate}
\usepackage{diagbox}
\usepackage{mathtools}
\usepackage{paralist}
\usepackage{tabularx}
\usepackage{lipsum}
\usepackage{ragged2e}
\usepackage{makecell}

\usepackage[T1]{fontenc}

\usepackage[utf8]{inputenc}

\usepackage{microtype}

\usepackage{inconsolata}

\newcommand{\name}{\textsc{ConKE}}

\title{\name: Conceptualization-Augmented Knowledge Editing in Large Language Models for Commonsense Reasoning}

\author{Liyu Zhang\thanks{Equal Contribution},
Weiqi Wang$^{*}$, 
Tianqing Fang,
Yangqiu Song\\
Department of Computer Science and Engineering, HKUST, Hong Kong SAR, China\\
\texttt{lzhangcx@connect.ust.hk, \{wwangbw, tfangaa, yqsong\}@cse.ust.hk}\\
}

\begin{document}
\maketitle
\begin{abstract}
Knowledge Editing (KE) aims to adjust a Large Language Model’s (LLM) internal representations and parameters to correct inaccuracies and improve output consistency without incurring the computational expense of re-training the entire model.
However, editing commonsense knowledge still faces difficulties, including limited knowledge coverage in existing resources, the infeasibility of annotating labels for an overabundance of commonsense knowledge, and the strict knowledge formats of current editing methods.
In this paper, we address these challenges by presenting \name, a framework that integrates conceptualization and instantiation into the KE pipeline for LLMs to enhance their commonsense reasoning capabilities.
\name{} dynamically diagnoses implausible commonsense knowledge within an LLM using another verifier LLM and augments the source knowledge to be edited with conceptualization for stronger generalizability.
Experimental results demonstrate that LLMs enhanced with \name{} successfully generate commonsense knowledge with improved plausibility compared to other baselines and achieve stronger performance across multiple question answering benchmarks.
Our data, code, and models are publicly available at \href{https://github.com/HKUST-KnowComp/ConKE}{https://github.com/HKUST-KnowComp/ConKE}.
\end{abstract}


\section{Introduction}
\label{sec:introduction}
Recent advancements in Large Language Models (LLMs;~\citealp{GPT4o,GPT4omini,LLAMA3,DBLP:conf/eacl/ChanCWJFLS24}) have led to Knowledge Editing (KE;~\citealp{survey1,survey2,lau2024ecomeditautomatedecommerceknowledge}), a computationally efficient strategy to correct inaccurate responses and update LLMs by modifying their internal weights or representations, without re-training the entire model.
Such methods have been applied to various domains, including factual reasoning~\cite{DBLP:conf/acl/JuCY0DZL24,DBLP:conf/acl/WangLSCXM24}, medical knowledge~\cite{DBLP:conf/cikm/XuZZLL00WY0C024}, and commonsense reasoning~\cite{DBLP:conf/emnlp/HuangW0024}, and have proven effective in enhancing domain-specific expertise.


Despite their success, current KE methods face several challenges, including limited knowledge coverage~\cite{CommonsenseReasoninginAI} in existing commonsense knowledge bases~\cite{DBLP:conf/emnlp/West0SLJLCHBB023,DBLP:conf/www/FangZWSH21,DBLP:conf/eacl/YangDCC23,DBLP:conf/emnlp/FangWCHZSH21,DBLP:journals/corr/abs-2304-10392,DBLP:conf/emnlp/0001WKLFBLYLLYY24,DBLP:conf/emnlp/Xu0S0JFBLYLLYYC24} which offer limited coverage and focus on isolated facts, rather than forming hierarchical structures that enable generalization through editing~\cite{DBLP:conf/emnlp/MaIFONO21,DBLP:journals/corr/abs-2406-10885}.
Furthermore, the unstructured nature of commonsense knowledge complicates scaling, while the flexible representation of commonsense knowledge means that a single fact may manifest in multiple formats. This necessitates editing at the \texttt{(relation, tail)} pair level rather than at individual tokens.

\begin{figure*}[t]
    \centering
    \includegraphics[width=\textwidth]{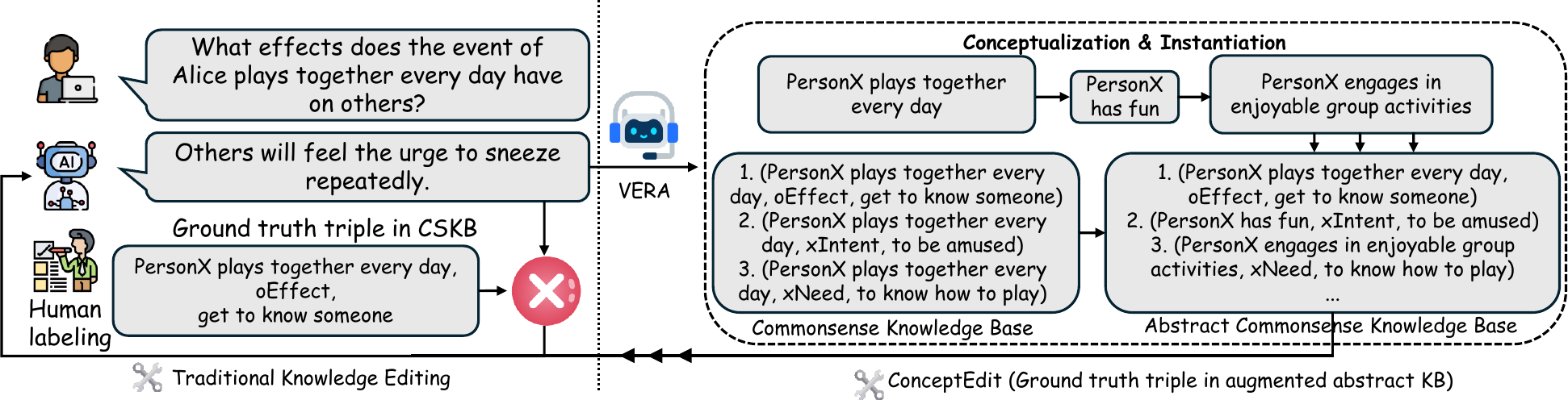}
    \caption{An overview of~\name{}, which pipelines conceptualization and instantiation, knowledge editing, and LLM verification together for automated and scalable knowledge editing over commonsense knowledge.}
    \label{fig:framework}
\end{figure*}

To address these issues, we present \name{}, a novel knowledge editing framework tailored for editing commonsense knowledge within LLMs.
We use VERA~\cite{Vera}, an automated commonsense plausibility verifier, to assess the plausibility of commonsense knowledge in LLMs.
For knowledge deemed erroneous and requiring edits, we integrate conceptualization and instantiation~\cite{CAT,CAR} to enrich semantic coverage and support more generalizable editing, covering not only the targeted knowledge but also other potentially relevant yet implausible information within the LLM. implausible information within the LLM. This pipeline therefore integrates automated detection, semantic enrichment, and edit application into one closed loop, enabling fully end-to-end scalability.

To ensure flexibility, \name{} adopts an open-ended format for editing, enabling the handling of arbitrary knowledge structures rather than focusing solely on traditional \texttt{(h,r,t)} triplets.
We go beyond traditional Knowledge Editing techniques by combining automated knowledge detection, conceptualization, and instantiation, enhancing the model's ability to generalize and adapt to diverse contexts.
Experimental results on AbstractATOMIC~\cite{abstractATM} demonstrate that LLMs enhanced by \name{} generate commonsense knowledge with improved plausibility.
Further evaluations across five commonsense question-answering benchmarks also show performance improvements. These experiments demonstrate the robustness and generalizability of our approach in enhancing commonsense reasoning across diverse architectures and tasks.

\section{Related Works}
\label{sec:related}
\subsection{Knowledge Editing}
Knowledge editing~\cite{DBLP:conf/emnlp/CaoAT21} aims to update an LLM’s internal knowledge without full retraining or relying solely on prompt engineering, is becoming increasingly crucial. 
\citet{ROME} propose ROME, which identifies and updates factual associations within specific MLP layers, achieving precise single-fact edits guided by causal mediation analysis.
MEMIT~\cite{MEMIT} extends ROME’s principles to handle large-scale edits simultaneously. By distributing updates across multiple layers and parameters, MEMIT efficiently integrates thousands of facts while maintaining specificity and fluency.
GRACE~\cite{GRACE}, on the other hand, avoids internal parameter changes by integrating external dictionaries and adapters as a modular memory source. 
This approach allows flexible, inference-time access to new knowledge, though it may sacrifice some internal coherence and interpretability.
In our work, we build upon these methods to enhance editing commonsense knowledge in LLMs.

\subsection{Conceptualization in Commonsense}
Conceptualization abstracts entities or events into general concepts, forming abstract commonsense knowledge~\cite{murphy2004big}, while instantiation grounds these concepts into new instances, introducing additional commonsense knowledge. 
Previous work largely focused on entity-level conceptualization~\cite{DBLP:conf/eacl/DurmeMS09,DBLP:conf/ijcai/SongWWLC11,DBLP:conf/ijcai/SongWW15,DBLP:journals/ai/LiuCWLCXCJ22,DBLP:conf/emnlp/PengWHJ0L0022}, with~\citet{abstractATM,CAT,CAR} pioneering event-level conceptualization from WordNet~\cite{DBLP:journals/cacm/Miller95} and Probase~\cite{DBLP:conf/sigmod/WuLWZ12}. 
For instantiation,~\citet{DBLP:conf/eacl/AllawayHBMDC23} introduced a controllable generative framework that automatically identifies valid instances.
In this work, we leverage the conceptualization distillation framework proposed by~\citet{CANDLE} to augment the knowledge being edited, ensuring broader semantic coverage and thereby improving the generalizability of edited knowledge.


\section{The \name{} Framework}
\label{sec:methodology}
An overview of~\name{} is presented in Figure~\ref{fig:framework}.
Our framework consists of three main components: (1) automated knowledge verification with VERA~\cite{Vera}, (2) abstract knowledge acquisition via conceptualization and instantiation, and (3) LLM knowledge editing.
We use the AbstractATOMIC~\cite{abstractATM} and CANDLE~\cite{CANDLE} datasets for training and evaluation as two rich sources of abstract knowledge with conceptualization and instantiation.
The training set of both datasets are used for editing and the testing sets are used for evaluation.

\subsection{Automated Knowledge Verification}
Since commonsense knowledge is vast, traditional human-in-the-loop methods for detecting and correcting erroneous outputs in LLMs are neither easily scalable nor adaptable.
Inspired by recent advances in using LLMs as automated judges~\cite{DBLP:conf/emnlp/RainaLG24,EcomScript}, we propose a fully automated verification strategy to assess an LLM’s internal commonsense knowledge.
Our verification process involves VERA~\cite{Vera}, a discriminative model trained to score the plausibility of arbitrary commonsense statements, as our evaluation tool.
For each triple in the AbstractATOMIC~\cite{abstractATM} training set, we prompt the LLM with the head event and request it to generate the corresponding relation and tail.
VERA then evaluates the plausibility of the generated knowledge by producing a score in the range $[0, 1]$, where values above 0.5 are considered plausible, and those below 0.5 are deemed implausible.
By iterating over all triples, this process provides both the LLM’s generated responses and VERA’s discrimination results, pinpointing which portions of the generated knowledge are incorrect.
Consequently, we can identify the exact ''areas'' within the LLM’s internal knowledge that require editing. 
This automated pipeline eliminates the dependence on costly human annotations for error detection, enabling scalable and efficient improvements of the LLM’s commonsense understanding.

\subsection{Conceptualization and Instantiation}
While existing approaches primarily integrate decontextualized commonsense knowledge into LLMs through KE techniques, we hypothesize that capturing the diverse patterns that the same piece of knowledge can exhibit under different contexts is equally important. However, repeated editing may result in knowledge drift, where successive modifications will lead to subtle conflicts, causing the model's internal representation to become unstable.
To this end, we augment the knowledge to be edited by implementing both conceptualization and instantiation, following~\citet{CANDLE}.
For each triple targeted for editing, we first abstract its instances into more general concepts by prompting GPT-4o, producing abstract knowledge triples (Figure~\ref{fig:framework}).
We then instantiate these abstract concepts into novel, context-specific instances, again using GPT-4o, thereby forming a rich knowledge base.
This process yields approximately 160,000 commonsense knowledge triples, substantially improving the semantic coverage and contextual adaptability of the edited knowledge.
Additionally, we are mindful of cascading effects that may arise when modifying a piece of commonsense knowledge. As noted in~\cite{survey2}, knowledge is highly interconnected, and modifying one fact can trigger unintended changes in related facts, leading to inconsistencies. To mitigate these cascading effects, we use conceptualization and instantiation to ensure that modifications to abstract concepts are consistently applied to their related instances, hence maintaining coherence and reducing the risk of introducing inconsistencies.

\subsection{LLM Knowledge Editing}
Finally, we apply knowledge editing to the LLM using the enriched knowledge base generated through our conceptualization and instantiation processes, correcting errors identified by VERA. 
To accomplish this, we experiment with three established knowledge editing methods: MEMIT~\cite{MEMIT}, ROME~\cite{ROME}, and GRACE~\cite{GRACE}. 
For GRACE, which relies on adapters to determine whether and how to use an external dictionary, we adopt the original deferral mechanism implementation.
We evaluate our framework with these editing methods on four representative LLM backbones: \texttt{Mistral-7B-Instruct-v0.2}\cite{Mistral}, \texttt{Meta-Llama-3-8B-Instruct}\cite{LLAMA3}, \texttt{Chatglm2-6b}\cite{DBLP:journals/corr/abs-2406-12793}, and \texttt{GPT-J-6B}\cite{gpt-j}.

\section{Experiments and Analyses}
In this section, we evaluate LLMs after applying~\name{} through expert and automated assessments, illustrating improved performance on downstream tasks and present several ablation studies.
\begin{figure}[t]
    \centering
    \includegraphics[width=1\linewidth]{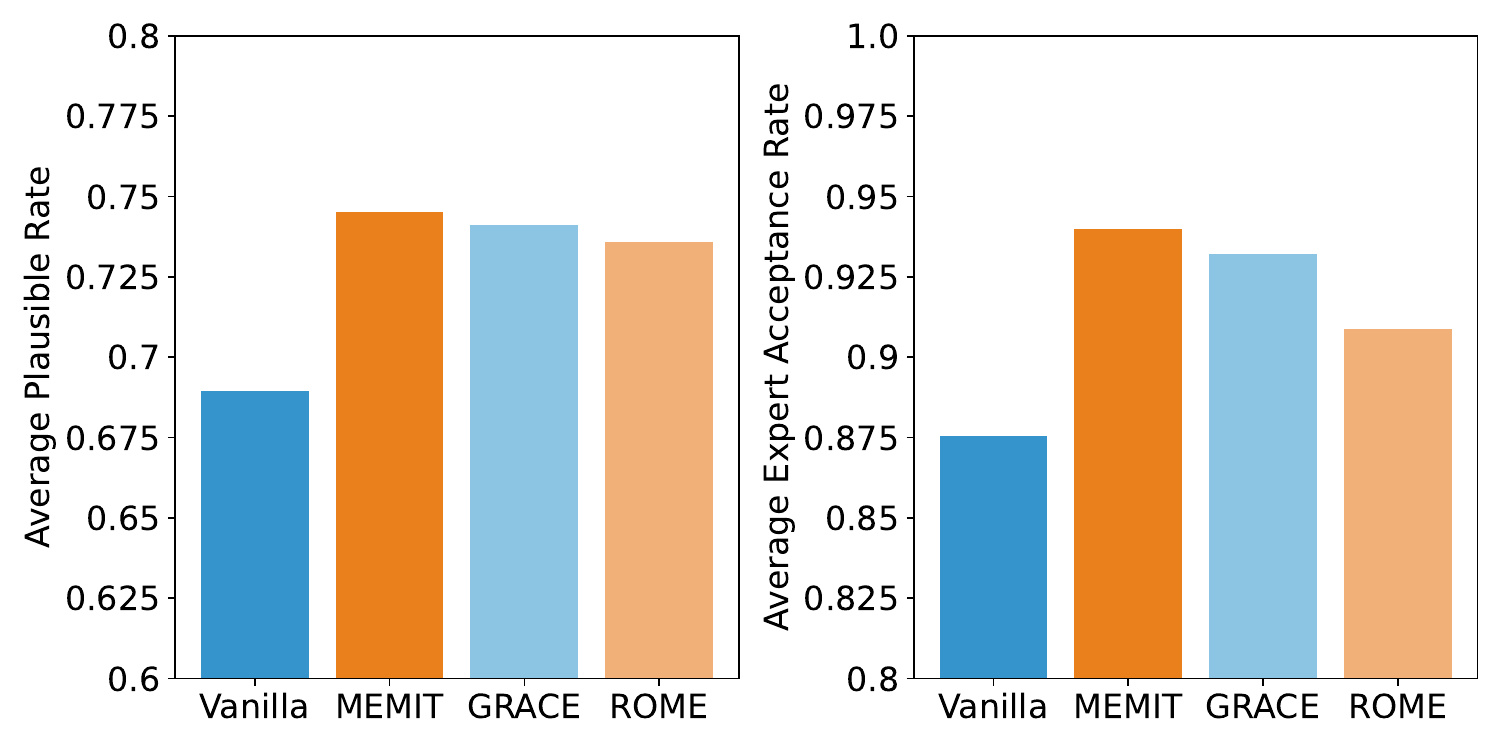}
    \caption{Average plausible rate and expert acceptance rate of LLMs' generation after~\name{}.}
    \label{fig:expert_result}
\end{figure}

\subsection{LLMs-After-Editing Evaluation}
We first evaluate LLMs after editing via two measures.
First, we prompt these LLMs with head events in the testing set of AbstractATOMIC and ask it to complete the commonsense knowledge.
With the generations on the testing set, we ask VERA to score them again and we calculate the plausible ratio whose scores are above 0.5.
Then, we sample a subset of 200 generations and recruit two expert annotators to conduct a manual analyses on the acceptance ratio of the plausible assertions that passed VERA's filtering.
We compare models after being edited with MEMIT, GRACE, and ROME, and set another vanilla group as baseline comparison.
As shown in Figure~\ref{fig:expert_result}, both VERA and human evaluations exhibit consistent trends, with
 human raters tend to assign higher scores but identifying similar improvements. 
When applying MEMIT-based editing, both VERA and human evaluations show notable enhancements over the Vanilla baseline. 
Similarly, GRACE and ROME edits enhance plausibility scores, with MEMIT and GRACE achieving the highest overall performance. 
The strong results from expert annotations further validate the reliability of VERA’s judgments, supporting the use of VERA in our framework as an effective commonsense evaluator to identify implausible knowledge requiring further editing. 
This approach reduces reliance on manual annotations while preserving robust assessment capabilities.

\begin{figure}[t]
    \centering
    \includegraphics[width=1\linewidth]{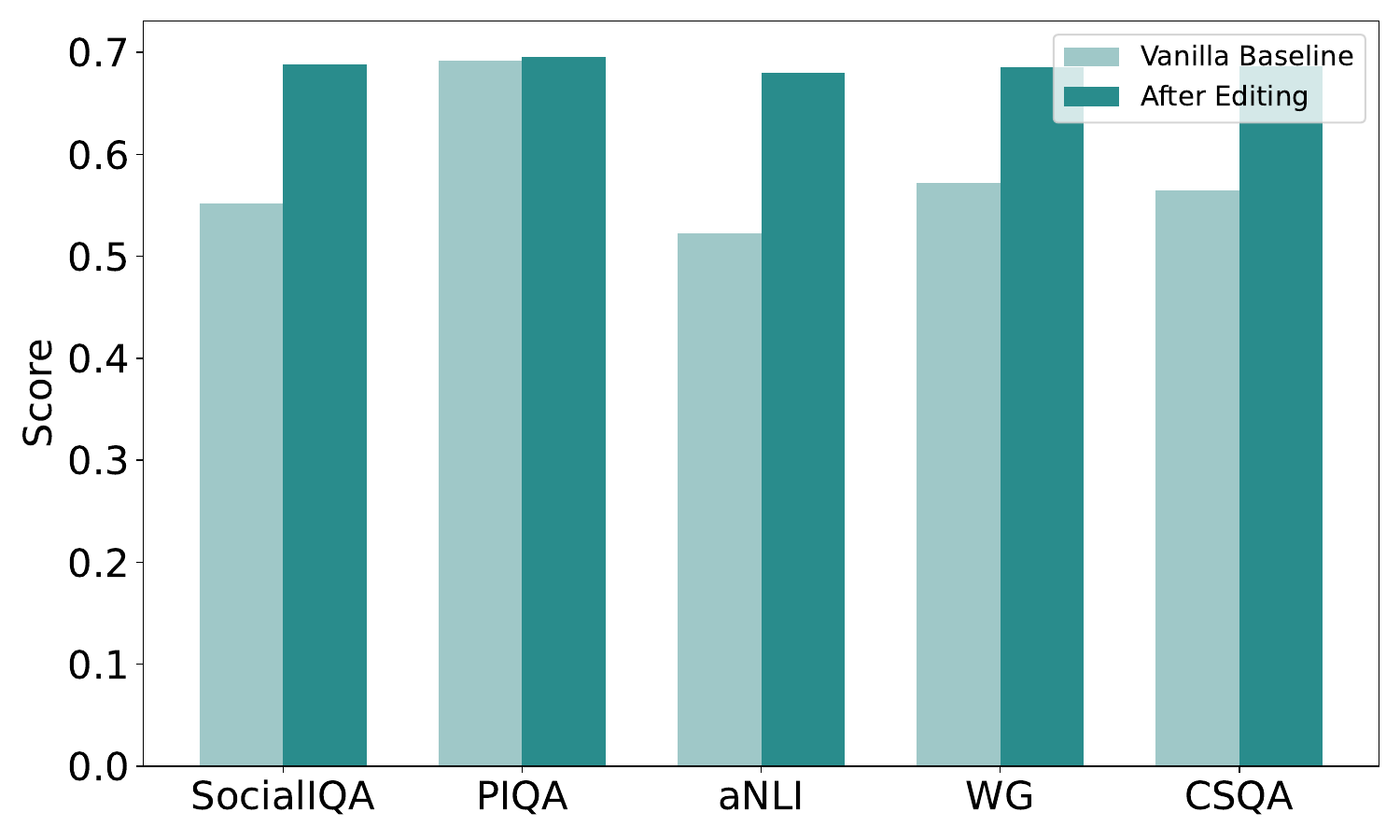}
    \caption{Performance of the best LLM after editing on five downstream tasks compared to the vanilla baseline.}
    \label{fig:benchmark_result}
\end{figure}

\subsection{Downstream Improvements}
To assess whether enhanced internal commonsense reasoning improves downstream task performance, we evaluate the edited models on multiple commonsense reasoning benchmarks. 
Following~\citet{DBLP:conf/aaai/MaIFBNO21}, we test our framework on the validation splits of five widely-used commonsense QA benchmarks: Abductive NLI (aNLI;~\citealp{DBLP:conf/iclr/BhagavatulaBMSH20}), CommonsenseQA (CSQA;~\citealp{DBLP:conf/naacl/TalmorHLB19}), PhysicalIQA (PIQA;~\citealp{DBLP:conf/aaai/BiskZLGC20}), SocialIQA (SocialIQA;~\citealp{DBLP:conf/emnlp/SapRCBC19}), and WinoGrande (WG;~\citealp{DBLP:journals/cacm/SakaguchiBBC21}). 
These benchmarks are designed to evaluate a range of knowledge types crucial for robust commonsense reasoning~\cite{DBLP:conf/emnlp/ShiWFX0LS23,DBLP:journals/corr/abs-2406-02106,shi2025inferencedynamicsefficientroutingllms,li2025patternsprinciplesfragilityinductive}.
We compare the performance of the best LLM edited with \name{} against its corresponding vanilla baseline across all benchmarks, with the results visualized in Figure~\ref{fig:benchmark_result}. 
The results show that models edited with \name{} achieve significant performance improvements across all benchmarks, with particularly notable gains in aNLI and SocialIQA. 
These findings demonstrate the effectiveness of \name{} in enhancing commonsense reasoning capabilities and suggest its potential for broader applications in improving LLM performance on real-world reasoning tasks.

\begin{figure}
    \centering
    \includegraphics[width=1\linewidth]{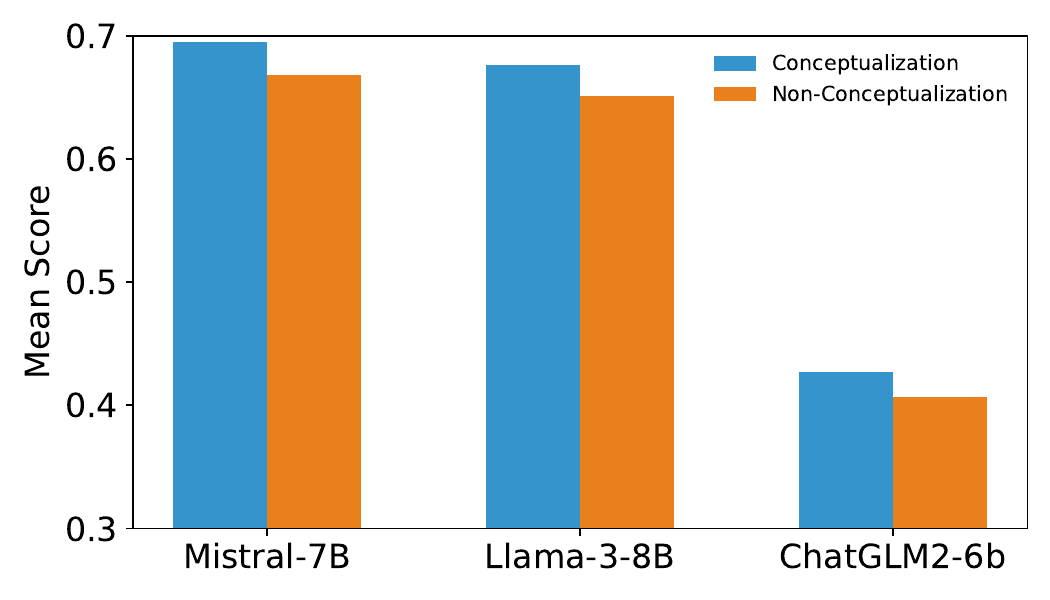}
    \caption{VERA evaluation scores of edited LLMs with and without integrating conceptualization.}
    \label{fig:ablation_result}
\end{figure}

\subsection{Ablation Study}
Finally, to validate the effect of conceptualization, we conducted an ablation study on MEMIT by removing the conceptualization step and comparing performance. 
In this setup, we edit LLMs both with and without the integration of conceptualization and instantiation, and evaluate their performance by examining the average VERA scores of the generated outputs on the testing set.
The conceptualized variant leveraged enriched commonsense triples generated via abstraction and instantiation prior to the editing process, while the non-conceptualized variant directly applied MEMIT without these pre-processing steps.

Figure~\ref{fig:ablation_result} demonstrates that the conceptualized variants consistently outperform their non-conceptualized counterparts, achieving higher plausibility and improved downstream task accuracy.
These results suggest that the enriched conceptual patterns introduced before editing not only enhance plausibility but also enable the model to generalize commonsense knowledge to more complex reasoning tasks, ultimately boosting overall performance~\cite{xu2025multiagentreasoningsystemscollaborative}.

\section{Conclusions}
In this paper, we introduce~\name{}, a novel knowledge editing framework designed to enhance commonsense reasoning in LLMs by addressing challenges of limited knowledge coverage and scalability, and by integrating automated verification through VERA and semantic enrichment via conceptualization and instantiation for more effective and generalizable editing. 
Experimental results demonstrate significant improvements in both knowledge plausibility and downstream task performance, validating the effectiveness of our approach.
We envision that \name{} will inspire future research on scalable and context-aware knowledge editing, paving the way for LLMs to better handle the complexity and diversity of commonsense reasoning.

\section*{Limitations}
Our approach, \name{}, advances LLM commonsense reasoning through conceptualization and iterative knowledge editing, yet several challenges persist. 
First, editing one piece of knowledge can cascade through related concepts, creating non-linear interactions that are difficult to detect and manage, especially as the knowledge base scales up. 
Second, iterative updates risk knowledge drift, where successive edits subtly conflict with or overwrite prior facts, emphasizing the need for robust frameworks to maintain consistency. 
Finally, the lack of stable ground truth for commonsense, which is often context-sensitive and culturally variable, complicates standardization. 
Addressing these challenges will require globally coordinated editing mechanisms, improved theoretical frameworks, and systematic human-in-the-loop validation to ensure edits align with broader consensus and expert judgment.

\section*{Ethics Statement}
In this paper, all datasets and models used are free and accessible for research purposes, aligning with their intended usage. 
The expert annotators are graduate students with extensive experience in NLP and commonsense reasoning research, and they voluntarily agreed to participate without compensation. 
\\ However, we recognize commonsense knowledge is inherently culturally and contextually variable, and there are ethical considerations related to the knowledge edited and propagated through the models. We must ensure that the knowledge inserted into the model doesn't favor certain views over others, especially in sensitive cases such as healthcare or law applications. To mitigate this, we implement a robust process of cross-validation with human experts to monitor for biases. Moreover, we propose regular audits of the system's performance, to ensure that its performance remains fair.

\section*{Acknowledgements}
We thank the anonymous reviewers and the area chair for their constructive comments. 
The authors of this paper were supported by the ITSP Platform Research Project (ITS/189/23FP) from the ITC of Hong Kong, SAR, China, as well as the AoE (AoE/E-601/24-N), the RIF (R6021-20), and the GRF (16205322) from the RGC of Hong Kong, SAR, China.

\bibliography{custom}




\end{document}